\documentclass{article}

\usepackage{fullpage}

\usepackage{natbib}

\usepackage[utf8]{inputenc} 
\usepackage[T1]{fontenc}    
\usepackage{hyperref}       
\usepackage{url}            
\usepackage{booktabs}       
\usepackage{amsfonts}       
\usepackage{nicefrac}       
\usepackage{microtype}      

\usepackage{amsmath, amsthm, amssymb}
\usepackage[parfill]{parskip}
\usepackage{amsfonts}
\usepackage{color}

\newtheorem{thm}{Theorem}[section]
\newtheorem{lem}{Lemma}[section]

\newcommand{\diag}{\mathrm{diag}}
\newcommand{\bbR}{\mathbb{R}}

\DeclareMathOperator*{\mini}{minimize}

\newcommand{\rank}{\mathrm{rank}}

\title{Depth Creates No Bad Local Minima}

\author{
\begin{tabular}[t]{c}
 Haihao Lu\\
Massachusetts Institute of Technology \\
\fontsize{11pt}{11pt}\texttt{haihao@mit.edu} \\
\end{tabular}
\begin{tabular}[t]{c}
Kenji Kawaguchi\\
Massachusetts Institute of Technology  \\
\fontsize{11pt}{11pt}\texttt{kawaguch@mit.edu} \\
\end{tabular}
}
\date{}

\begin{document}

\maketitle

\begin{abstract}
  In deep learning, \textit{depth}, as well as \textit{nonlinearity}, create non-convex loss surfaces. Then, does depth alone create bad local minima? In this paper, we prove that without nonlinearity, depth alone does not create bad local minima, although it induces non-convex loss surface. Using this  insight, we greatly simplify a recently proposed proof to show that all of the local minima of feedforward deep linear neural networks are global minima. Our theoretical results generalize previous results with fewer assumptions, and this analysis provides a method to show similar results beyond square loss in deep linear models.
\end{abstract}

\section{Introduction}
Deep learning has recently had a profound impact on
the machine learning, computer vision, and artificial intelligence
communities. In addition to its practical successes, previous studies have revealed several reasons why deep learning has been successful from the viewpoint of its \textit{model classes}. An (over-)simplified explanation is the harmony of its great expressivity and \textit{big data}: because of its great expressivity, deep learning can have less \textit{bias}, while a large training dataset leads to less \textit{variance}. The great expressivity can be seen from an aspect of representation learning as well:
whereas traditional machine learning makes use of features designed by human users or experts as a type of prior, deep learning tries to learn features from the data as well. More accurately, a key aspect of the model classes in deep learning is the \textit{generalization} property; despite its great expressivity, deep learning model classes can maintain great generalization properties \citep{livni2014computational,mhaskar2016learning,poggio2016and}. This would distinguish deep learning from other possibly too flexible methods, such as shallow neural networks with too many hidden units, and traditional kernel methods with a too powerful kernel. Therefore, the practical success of deep learning seems to be supported by the great quality of its model classes.   

However, having a great model class is not so useful if we cannot find a good model in the model class via training. Training a deep model is typically framed as non-convex optimization. Because of its non-convexity and high dimensionality, it has been unclear whether we can \textit{efficiently} train a deep model. 
Note that the difficulty comes from the combination of non-convexity and high dimensionality in weight parameters. If we can reformulate the training problem into several decoupled training problems, with each having a small number of weight parameters, we can effectively train a model via non-convex optimization as theoretically shown in Bayesian optimization and global optimization literatures \citep{kawaguchiNIPS2015,wang2016optimization,kawaguchi2016global}. As a result of non-convexity and high-dimensionality, it was shown that training a general neural network model is NP-hard \citep{blum1992training}. However, such a hardness-result in a worst case analysis  would not tightly capture what is going on in practice, as we seem to be able to efficiently train deep models in practice. 

To understand its practical success beyond worst case analysis, theoretical and practical investigations on the training of deep models have recently become an active research area \citep{saxe2013exact,dauphin2014identifying,choromanska2015loss,haeffele2015global,shamir2016distribution,kawaguchi2016deep,swirszcz2016local,arora2016understanding,freeman2016topology,soudry2017exponentially}.

An important property of a deep model is that the non-convexity comes from \textit{depth}, as well as \textit{nonlinearity}: indeed, depth by itself creates highly non-convex optimization problems. One way to see a property of the non-convexity induced by depth is the non-uniqueness owing to \textit{weight--space symmetries} \citep{kuurkova1994functionally}: the  model represents the same function mapping from the input to the output with different distinct settings in the weight space.  Accordingly, there are many distinct globally optimal points and many distinct points with the same loss values due to weight--space symmetries, which would  result in a non-convex epigraph (i.e., non-convex function) as well as non-convex sublevel sets (i.e., non-quasiconvex function). Thus, it has been unclear whether \textit{depth} by itself can create a difficult non-convex loss surface. The recent  work \citep{kawaguchi2016deep} indirectly showed, as a consequence of its main theoretical results, that depth does not create bad local minima of deep linear model with Frobenius norm although it creates potentially bad saddle points. 

In this paper, we directly prove that all local minima of deep linear model corresponds to local minima of shallow model. Building upon this new theoretical insight, we propose a simpler proof for one of the main results in the recent work \citep{kawaguchi2016deep}; all of the local minima of feedforward deep linear neural networks with Frobenius norm are global minima. The power of this proof can go beyond Frobenius norm: as long as the loss function satisfies Theorem \ref{thm:rank}, all local minima of deep linear model corresponds to local minimum of shallow model.      

%
%
%
%
%

\section{Main Result}

To examine the effect of depth alone, we consider the following optimization problem of feedforward deep linear neural networks with the square error loss: 
\begin{align}{\label{eq:obj}}
\mini_{W} \;\  L(W)=\frac{1}{2}\|W_{H} W_{H-1}\cdots W_{1}X-Y\|_{F}^{2},
\end{align}

where $W_{i}\in \bbR^{d_{i}\times d_{i-1}}$ is the weight matrix,  $X\in \bbR^{d_0\times m}$ is the input training data, and $Y\in \bbR^{d_H\times m}$ is the target training data. Let $p=\arg\min_{0\le i\le H}d_{i}$ be the index corresponding to the smallest width. Note that for any $W$, we have $\rank (W_{H} W_{H-1}\cdots W_{1}) \le d_p$. To analyze optimization problem (\ref{eq:obj}), we also consider the following optimization problem with a ``shallow'' linear model, which is equivalent to problem (\ref{eq:obj}) in terms of the global minimum value: 
\begin{align}\label{eq:newprob}
\mini_{R} \;\  F(R)= \|RX-Y\|_{F}^{2} \;\ \ \ \text{ s.t. }  \ \ \;\  \rank(R)\le d_p,
\end{align}
where $R\in \bbR^{d_H\times d_0}$. Note that problem (\ref{eq:newprob}) is non-convex, unless \(d_p = \min(d_H,d_0)\), whereas problem (\ref{eq:obj}) is non-convex, even when \(d_p \ge \min(d_H,d_0)\) with \(H>1\). In other words, deep parameterization creates a non-convex loss surface even without nonlinearity. 

Though we only consider the Frobenius loss here, the proof holds for general cases. As long as the loss function satisfies Theorem \ref{thm:rank}, all local minima of deep linear model corresponds to local minimum of shallow model.

Our first main result states that even though deep parameterization creates a non-convex loss surface, it does not create new bad local minima. In other words, every local minimum in problem (\ref{eq:obj}) corresponds to a local minimum in problem (\ref{eq:newprob}).

\vspace{6pt}
\begin{thm} \label{thm:main1}
\emph{(Depth creates no new bad local minima)}
Assume that $X$ and $Y$ have full row rank. If $\bar W=\{\bar W_{1},\dots,\bar W_{H}\}$ is a local minimum of problem (\ref{eq:obj}), then $\bar R = \bar W_{H} \bar W_{H-1}\cdots \bar W_{1}$ achieves the value of a local minimum of problem (\ref{eq:newprob}).
\end{thm} 
\vspace{6pt}

Therefore, we can deduce the property of the local minima in problem (\ref{eq:obj}) from those in problem (\ref{eq:newprob}). Accordingly, we first analyze the local minima in problem (\ref{eq:newprob}), and obtain the following statement.  
\vspace{6pt}  
\begin{thm}  \label{thm:main2}
\emph{(No bad local minima for rank restricted shallow model)}
If $X$ has full row rank, all local minima of optimization problem \eqref{eq:newprob} are global minima.
\end{thm}
\vspace{6pt}  

By combining Theorems \ref{thm:main1} and \ref{thm:main2}, we conclude that every local minimum is a global minimum for feedforward deep linear networks with a square error loss.   

\vspace{6pt}  
\begin{thm} \label{thm:main3}
\emph{(No bad local minima for deep linear neural networks)}
If $X$ and $Y$ have full row rank, then all local minima of problem (\ref{eq:obj}) are global minima.
\end{thm} 
\vspace{6pt}  

Theorem \ref{thm:main3} generalizes one of the main results in  \citep{kawaguchi2016deep} with fewer assumptions. Following the theoretical work with a random matrix theory \citep{dauphin2014identifying,choromanska2015loss}, the recent work \citep{kawaguchi2016deep}  showed that under some strong assumptions, all of the local minima are global minima for a class of nonlinear deep networks.
Furthermore, the recent work \citep{kawaguchi2016deep} proved the following properties for a class of general deep linear networks with arbitrary depth and width: 1) the objective function is non-convex and non-concave; 
2) all of the local minima are global minima;
3) every other critical point is a saddle point; and
4) there is no saddle point with the Hessian having no negative eigenvalue for shallow networks with one hidden layer, whereas such saddle points exist for deeper networks. Theorem \ref{thm:main3} generalizes the second statement  with fewer assumptions; the  previous papers \citep{baldi1989linear,kawaguchi2016deep} assume that the data matrix $YX^T(XX^T)^{-1}XY^T$ has distinct eigenvalues, whereas we do not assume that.  

\section{Proof}
In this section, we provide the proofs of Theorems \ref{thm:main1}, \ref{thm:main2}, and \ref{thm:main3}.

\subsection{Proof of Theorem \ref{thm:main1}}
In order to deduce the proof of Theorem \ref{thm:main1}, we need some fundamental facts in linear algebra. The next two lemmas recall some basic facts of perturbation theory for singular value decomposition (SVD).

Let $M$ and $\bar{M}$ be two $m\times n$ ($m\ge n$) matrices with
SVDs
\[
B=U\Sigma V^{T}=(U_{1},U_{2})\left(\begin{array}{cc}
\Sigma_{1}\\
 & \Sigma_{2}\\
\\
\end{array}\right)\left(\begin{array}{c}
V_{1}^{T}\\
V_{2}^{T}
\end{array}\right)
\]

\[
\bar{B}=\bar{U}\bar{\Sigma}\bar{V}^{T}=(\bar{U}_{1},\bar{U_{2}})\left(\begin{array}{cc}
\bar{\Sigma}_{1}\\
 & \bar{\Sigma}_{2}\\
\\
\end{array}\right)\left(\begin{array}{c}
\bar{V}_{1}^{T}\\
\bar{V}_{2}^{T}
\end{array}\right),
\]
where $\Sigma_{1}=\diag(\sigma_{1},\cdots,\sigma_{k})$, $\Sigma_{2}=\diag(\sigma_{k+1},\cdots,\sigma_{n})$,
$\Sigma_{1}=\diag(\bar{\sigma}_{1},\cdots,\bar{\sigma}_{k})$, $\Sigma_{2}=\diag(\bar{\sigma}_{k+1},\cdots,\bar{\sigma}_{n})$, $U$, $V$, $\bar{U}$ and $\bar{V}$ are orthogonal matrices.

\vspace{6pt}  
\begin{lem} {\bf Continuity of Singular Value}\label{thm: singular value}
The singular value $\sigma_i$ of a matrix is a continuous map of entries of the matrix.
\end{lem}
\vspace{6pt}  

\begin{lem} {\bf\citep{wedin1972perturbation} Continuity of Singular Space}\label{thm: Wedin}

If $$\rho:=\min\left\{ \min_{1\le i\le k,1\le j\le n-k}|\sigma_{i}-\bar{\sigma}_{k+j}|,\min_{1\le i\le k}\sigma_{i}\right\}>0 ,$$
then:
\begin{align*}
 \sqrt{\|\sin(U_{1},\bar{U}_{1})\|_{F}^{2}+\|\sin(V_{1},\bar{V}_{1})\|_{F}^{2}}
\le\frac{\sqrt{\|\left(\bar{M}-M\right)V_{1}\|_{F}^{2}+\|\left(\bar{M}^{*}-M^{*}\right)U_{1}\|_{F}^{2}}}{\rho}.
\end{align*}
\end{lem}

For a fixed matrix $B$, we say ``matrix $A$ is a perturbation of matrix $B$'' if $\|A - B\|_{\infty}$ is $o(1)$, which means that the difference between $A$ and $B$ is much smaller than any non-zero number in matrix $B$.

Lemma \ref{thm: Wedin} implies that any SVD for a perturbed matrix is a perturbation of some SVD for the original matrix under full rank condition. More formally:
\begin{lem}\label{thm: perturbsvd}
Let $\bar{M}$ be a full-rank matrix with singular value decomposition $\bar{M}=\bar{U}\bar{\Sigma}\bar{V}^{T}$. $M$ is a perturbation of $\bar{M}$. Then, there exists one SVD of $M$, $M=U\Sigma V^{T}$, such that $U$ is a perturbation of $\bar{U}$, $\Sigma$ is a perturbation of $\bar{\Sigma}$ and $V$ is a perturbation of $\bar{V}$.(Notice that SVD  of a matrix may not be unique due to rotation of the eigen-space corresponding to the same eigenvalue)
\end{lem}
{\bf Proof:} With the small perturbation of matrix $\bar{M}$, Lemma \ref{thm: singular value} shows that the singular values does not change much. Thus, if $\|\bar{M}-M\|_{\infty}$ is small enough, $|\sigma_{i}-\bar{\sigma}_{i}|$ is also small for all \(i\). Remember that all singular values of $\bar{M}$ are positive. By letting $\Sigma_{1}$ contain only the singular value $\sigma_{i}$ (which may be multiple, and hence $U_{1}$ and $V_{1}$ are the singular spaces corresponding to the singular value $\sigma_{i}$), we have $\rho > 0$ in Lemma \ref{thm: Wedin}, thus Lemma \ref{thm: Wedin} implies that the singular space of the perturbed matrix corresponding to singular value $\sigma_{i}$ in the initial matrix does not change much. The statement of the lemma follows by combining this result for the different singular values together (i.e., consider each index \(i\) for different \(\sigma_i\) in the above argument).  \qed

We say that $W$ satisfies the rank condition, if $\rank(W_H\cdots    W_1) = d_p$. Any perturbation of the products of matrices is the product of the perturbed matrices, when the original matrix satisfies the rank constraint. More formally:

\vspace{6pt}  
\begin{thm}\label{thm:pertofmatrix}  Let $\bar R=\bar W_{H} \bar W_{H-1}\cdots \bar W_{1}$ with \(\rank(\bar R)=d_p\). Then, for any $R$, such that $R$ is a perturbation of $\bar{R}$ and $\rank(R)\le d_{p}$, there exists $\{W_1,W_{2},\dots,W_{H}\}$, such that $W_i$ is perturbation of $\bar{W}_i$ for all $i \in \{1,\ldots,H\}$ and $ R=W_{H} W_{H-1}\cdots W_{1}$.
\end{thm}
\vspace{6pt}  

We will prove the theorem by induction. When $H=2$, we can easily show that the perturbation of the  product of two matrices is the product of one matrix and the perturbation of the other matrix. When $H=k>=3$, we let $M$ be the product of two specific matrices, and by induction the perturbation of the product ($R$) is the product of a perturbation of $M$ and perturbations of the other $H-2$ matrix. And a perturbation of $M$ is also the product of perturbations of those two specific matrices, which proves the statement when $H=k$.

{\bf Proof:} The case with \(H=1\) holds by setting \(W_1=R\). We prove the lemma with \(H\ge 2\) by induction. 

We first consider the base case where $H=2$ with  $\bar R=\bar W_{2}\bar W_{1}$. 

Let  $\bar R=\bar U \bar \Sigma \bar V^T$ be the SVD of $\bar R$. It follows Lemma \ref{thm: perturbsvd} that there exists an SVD of $R$, $ R = U \Sigma V^T$, such that $U$ is a perturbation of $\bar{U}$, $\Sigma$ is a perturbation of $\bar{Sigma}$ and $V$ is a perturbation of $\bar{V}$. Because $\rank(\bar R)=d_{p}$, with a small perturbation, the positive singular values remain strictly positive, whereby, $\rank(R)\ge d_{p}$. Together with the assumption $\rank(R)\le d_{p}$, we have $\rank(R)=d_{p}$. Let $\bar S_2 = \bar U^T \bar W_2$ and $\bar S_1 = \bar W_1\bar V$. Note that $\bar U \bar \Sigma \bar V^T= \bar R= \bar W_{2} \bar W_{1}$. Hence, $\bar S_2 \bar S_1 = \bar \Sigma$ is a diagonal matrix. Remember $\Sigma$ is a perturbation of $\bar \Sigma$, thus there is an $S_2$, which is a perturbation of $\bar S_2$ (each row of $S_2$ is a scale of the corresponding row of $\bar S_2$), such that $S_2\bar S_1 = \Sigma$. Let $W_{2}=US_{2}$ and $W_{1}=\bar S_1 V$. Then, $W_1$ is a perturbation of $\bar W_1$, $W_2$ is a perturbation of $\bar W_2$, and $W_{1}W_{2}=R$, which proves the case when $H=2$.

For the inductive step, given that the lemma holds for the case with $H= k\ge 2$, let us consider the case when $H=k+1\ge 3$ with $\bar R=\bar W_{k+1} \bar W_{k} \cdots \bar W_{1}$. Let $\mathcal{I}$ be an index set defined as \(\mathcal I = \{p,p-1\}\) if \(p\ge 2\), \(\mathcal I = \{p+2,p+1\}\) if $p=0$ or $p=1$. We denote the \(i\)-th element of a set \(\mathcal I\) by $\mathcal I_i$. Then, $\bar M =\bar W_{\mathcal I_2} \bar W_{\mathcal I_1}$ exists as \(k+1\ge 3\). Note that $\bar R$ can be written as a product of $k$ matrices with \(\bar M\) (for example,  $\bar R=\bar W_{H} \cdots \bar W_{I_1+1}\bar M \bar W_{I_2-1}\cdots \bar W_{1}$). Thus, from the inductive hypothesis, for any $R$, such that $R$ is a perturbation of $\bar R$ and $\rank(R)\le d_{p}$, there exists a set of desired  $k$ matrices \(M\) and \(W_i\) for \(i\in \{1,\dots,k+1\} \setminus \mathcal I \), such that $W_i$ is perturbation of $\bar W_i$ for all $i\in \{1,\dots,k+1\} \setminus \mathcal I$,  $M$ is perturbation of $\bar M$,  and the product is equal to \(R\). Meanwhile, because \(\bar M\) is either a \(d_p\) by \(d_{p-2}\) matrix or a \(d_{p + 2}\) by \(d_p\) matrix, we have \(\rank(\bar M)\le d_p\) and $\rank(M) \le d_p$, and it follows \(\rank(\bar R)=d_p\) that \(\rank(\bar M)=d_p\). Thus, by setting \(\bar R \leftarrow \bar M\) and $R\ \leftarrow M$ (note that \(d_p\) in $\bar R = \bar W_{k+1} \bar W_{k} \cdots \bar W_1$ is equal to \(d_p\) in $\bar M =\bar W_{\mathcal I_2} \bar W_{\mathcal I_1} $), we can apply the proof for the case of $H=2$ to conclude: there exists $\{ W_{\mathcal I_2}, W_{\mathcal I_1}\}$, such that $W_i$ is perturbation of $\bar W_i$ for all \(i \in \mathcal I\), and $ M= W_{\mathcal I_2} W_{\mathcal I_1}$. Combined with the above statement from the inductive hypothesis, this implies the lemma with $H=k+1$, whereby we finish the proof by induction. 
\qed

The next two theorems show that, for any local minimum of $L(\cdot)$, there is another local minimum of $L(\cdot)$, whose function value is the same as the original and it satisfies the rank constraint.

\vspace{6pt}  
\begin{thm}\label{thm:rank} Let $W=\{W_{1},\cdots,W_{H}\}$ be a local minimum of
problem (\ref{eq:obj}) and $R\triangleq W_{H} W_{H-1}\cdots W_{1}$. If $W_i$ is not of full rank, then there
exists a $\bar{W_{i}}$, such that $\bar{W}_i$ is of full rank, $\bar W_i$ is a perturbation of $W_i$,  $\bar{W}=\{W_{1},\cdots,W_{i-1},\bar{W}_{i},W_{i+1},\cdots,W_{H}\}$
is a local minimum of problem (\ref{eq:obj}), and $L(W)=L(\bar{W})$. 
\end{thm}
\vspace{6pt}  

The idea of the proof is that if we just change one weight $W_i$ and keep all other weights, it becomes a convex least square problem. Then we are able to perturb $W_i$ to maintain the objective value as well as the perturbation is full rank.

{\bf Proof of Theorem \ref{thm:rank}} For notational convenience, let $A=W_{i-1}\cdots W_{1}X$
and $B=W_{i+1}\cdots W_{H}$, and let $L_{i}(W_{i})=\frac{1}{2}\|B^{T}W_{i}A-Y\|_{F}^{2}$. Because
$W$ is a local minimum of $L$, $W_{i}$ is a local minimum
of $L_{i}$. Let $A=U^T_{1}D_{1}V_{1}$ and $B=U^T_{2}D_{2}V_{2}$ are
the SVDs of $A$ and $B$, respectively, where $D_{i}$ is a diagonal
matrix with the first $s_{i}$ terms being strictly positive, $i=1,2$.
Minimizing $L_{i}$ over $W_{i}$ is a least square problem, and the
normal equation is

\begin{equation}\label{eq: normal_eq}
BB^{T}W_{i}AA^{T}=BYA^{T},
\end{equation}

hence
\begin{align*}
W_{i}&\in(BB^{T})^{+}BYA^{T}(AA^{T})^{+}+\left\{ M|BB^{T}MAA^{T}=0\right\} 
\\ & =U_{2}D_{2}^{+}V_{2}^{T}YV_{1}D_{1}^{+}U_{1}^{T}+\left\{ U_{2}KU_{1}^T|K_{1:s_{2},1:s_{1}}=0\right\}, 
\end{align*}
 
where $\left(\cdot\right)^{+}$ is a Moore\textendash Penrose pseudo-inverse
and $K$ is a matrix with suitable dimension with the entries in the
top left $s_{2}\times s_{1}$ rectangular being $0$. 

Since $V_{2}^{T}YV_{1}$ is of full rank, 
\begin{align*}
\rank(D_{2}^{+}V_{2}^{T}YV_{1}D_{1}^{+})
\ge\max\left\{ 0,s_{2}+s_{1}-\max\{d_{i},d_{i-1}\}\right\} \end{align*}
Thus, we can choose a proper $K$ (which contains $d_{i}+d_{i-1}-s_{2}-s_{1}$
$1$s at proper positions with all other terms being $0$s)  such that $D_{2}^{+}V_{2}^{T}YV_{1}D_{1}^{+}+K$
is of full rank, whereby $U_{2}\left(D_{2}^{+}V_{2}^{T}YV_{1}D_{1}^{+}+K\right)U_{1}^{T}$
is of full rank. Therefore, there is a full rank $\hat{W_{i}}$ that
satisfies the normal equation \eqref{eq: normal_eq}.

Let $\bar{W}_{i}(\mu)=W_i+ \mu\left(\hat{W}_{i}-W_i\right)$.
Then, $\bar{W}_{i}(\mu)$ also satisfies the normal equation, and  $L(\bar{W}(\mu))=L_{i}(\bar{W}_{i}(\mu))=L_{i}(W_{i})=L(W)$,
for any $\mu>0$.

Note that $W$ is a local minimum of $L(W)$. Thus, there exists a $\delta>0$,
such that for any $W^{0}$ satisfying $\|W^{0}-W\|_{\infty}\le\delta$, we have
$L(W^{0})\ge L(W)$. It follows from $\hat{W_{i}}$ being full rank that
there exists a small enough $\mu$, such that $\bar{W}_{i}(\mu)$
is full rank and $\|\bar{W}_{i}(\mu)-W_{i}\|_{\infty}$ is arbitrarily small (in particular, $\|\bar{W}_{i}(\mu)-W_{i}\|_{\infty}\le\frac{\delta}{2}$), because the non-full-rank matrices are discrete on the line of $\bar{W}_{i}(\mu)$
with parameter $\mu>0$ by considering the determine of $W_i^T(\mu)W_i(\mu)$ or  $W_i(\mu)W_i^T(\mu)$ as a polynomial of $\lambda$. Therefore, for any $W^{0}$, such that $\|W^{0}-\bar{W}(\mu)\|_{\infty}\le\frac{\delta}{2}$,
we have $$\|W^{0}-W\|_{\infty}\le\|W^{0}-\bar{W}(\mu)\|_{\infty}+\|\bar{W}_{i}(\mu)-W_{i}\|_{\infty}\le\delta \ ,$$
whereby $$L(W^0) \geq L(W) = L(\bar{W}(\mu)) \ .$$ This shows that $\bar{W}(\mu)=\left\{W_{1},\cdots,W_{i-1},\bar{W}_{i}(\mu),W_{i+1},\cdots,W_{H}\right\}$ is also a local minimum of problem (\ref{eq:obj}) for some small enough $\mu$. \qed


\vspace{6pt}  
\begin{lem}\label{lem:pert_two}
Let $R=AB$ for two given matrices $A\in R^{d_{1}\times d_{2}}$ and $B\in R^{d_{2}\times d_{3}}$. If $d_{1}\le d_{2}$, $d_{1}\le d_{3}$ and $rank(A)=d_{1}$, then any perturbation of $R$ is the product of $A$ and perturbation of $B$.
\end{lem}

{\bf Proof:} Let $A=UDV^{T}$ be the SVD of $A$, then, $R=UDV^{T}B$. Let
$\bar{R}$ be a perturbation of $R$ and let $\bar{B}=B+VD^{+}U^{T}(\bar{R}-R)$.
Then, $\bar{B}$ is a perturbation of $B$ and $A\bar{B}=\bar{R}$ by noticing $DD^{+}=I$, as $A$ has full row rank. \qed

\vspace{6pt}  
\begin{thm} \label{thm:rank2}
If $\bar{W}=\{\bar{W}_{1},\cdots,\bar{W}_{H}\}$ is a local
minimum with $\bar{W}_{i}$ being full rank, then, there exists $\hat{W}=\left\{ \hat{W}_{1},\cdots,\hat{W}_{H}\right\} $,
such that $\hat{W}_i$ is a perturbation of $\bar{W_i}$  for all
$i\in\left\{ 1,\ldots,H\right\} $, $\hat{W}$ is a local minimum,
$L(\hat{W})=L(\bar{W})$, and $\rank(\hat{W}_{H}\hat W_{H-1}\cdots\hat{W}_{1})=d_{p}$.
\end{thm}
\vspace{6pt}  

In the proof of Theorem \ref{thm:rank2}, we will use Theorem \ref{thm:rank} and Lemma \ref{lem:pert_two} to show that we can perturb $\bar{W}_{p-1}, \ \bar{W}_{p-2}\ldots, \bar{W}_1$ in sequence to make sure the perturbed weight is still the optimal solution and $\rank(\hat{W}_{p}\hat{W}_{p-1})=d_{p}$. Similar strategy can make sure $\rank(\hat{W}_{H}\hat{W}_{H-1}\cdots\hat{W}_{p+1})=d_{p}$, which then proves the whole theorem.

{\bf Proof of Theorem \ref{thm:rank2} :}  If $p\not=1$, consider $$L_1(T):=\|\bar{W}_{H}\cdots \bar{W}_{p+1}T \bar{W}_{p-2}\cdots \bar{W}_{1}X-Y\|_{F}^{2}.$$
Then, it follows from Lemma \ref{lem:pert_two} and $\bar{W}$ is a local minimum of $L(W)$ that $\bar{T}$ is a local minimum of $L_1$, where $\bar{T}=\bar{W}_{p}\bar{W}_{p-1}$.
It follows from Theorem \ref{thm:rank} that there exists $\hat{T}$, such
that $\hat{T}$ is close enough to $\bar{T}$, $\hat{T}$ is a local
minimum of $L_{1}(T)$, $L_{1}(\hat{T})=L_{1}(\bar{T})$, and $\rank(\hat{T})=d_{p}$.
Note $\hat{T}$ is a perturbation of $\bar{T}$, whereby, from Lemma \ref{lem:pert_two},
there exists $\hat{W}_{p}$, $\hat{W}_{p-1}$, which are perturbations
of $\bar{W}_{p}$ and $\bar{W}_{p-1}$, respectively, such that $\hat{W}_p\hat{W}_{p-1} = \hat{T}$. Thus, $\hat{W}^{0}=\left(\bar{W}_{H},\cdots,\bar{W}_{p+1},\hat{W}_{p},\hat{W}_{p-1},\bar{W}_{p-2}\cdots,\bar{W}_{1}\right)$
is a local minimum of $L(W)$, $L(\hat{W})=L(\bar{W})$ and $\rank(\hat{W}_{p}\hat{W}_{p-1})=d_{p}$. 

By that analogy, we can find $\hat{W}_{p}\cdots\hat{W}_{1}$, such
that $\hat{W}^{1}=\left(\bar{W}_{H},\cdots,\bar{W}_{p+1},\hat{W}_{p},\hat{W}_{p-1},\cdots,\hat{W}_{1}\right)$
is a local minimum of $L(W)$, $\hat{W}_{i}$ is a perturbation of
$\bar{W}_{i}$ for $i=1,\cdots,p$, $L(\hat{W}^{1})=L(\bar{W})$ and
$\rank(\hat{W}_{p}\hat{W}_{p-1}\cdots\hat{W}_{1})=d_{p}$. 

Similarly, we can find $\hat{W}_{H}\cdots\hat{W}_{p+1}$, such that $\hat{W}^{2}=\left(\hat{W}_{H},\cdots,\hat{W}_{p+1},\hat{W}_{p},\hat{W}_{p-1},\cdots,\hat{W}_{1}\right)$
is a local minimum of $L(W)$, $\hat{W}_{i}$ is a perturbation of
$\bar{W}_{i}$ for $i=p+1,\cdots,H$, $L(\hat{W}^{2})=L(\hat{W}^{1})=L(\bar{W})$
and $\rank(\hat{W}_{H}\hat{W}_{H-1}\cdots\hat{W}_{p+1})=d_{p}$. 

Noticing that 
\begin{align*}
\rank(\hat{W}_{H}\cdots\hat{W}_{1}) 
 \ge  \rank(\hat{W}_{H}\hat{W}_{H-1}\cdots\hat{W}_{p+1})
+\rank(\hat{W}_{p}\hat{W}_{p-1}\cdots\hat{W}_{1})-d_{p} = d_{p}
\end{align*}
and $\rank(\hat{W}_{H}\cdots\hat{W}_{1})\le\min_{i=0,\ldots,H}d_{i}=d_{p}$,
we have $\rank(\hat{W}_{H}\cdots\hat{W}_{1})=d_{p}$, which completes
the proof. \qed

\vspace{6pt}  
{\bf Proof of Theorem \ref{thm:main1}:}
It follows from Theorem \ref{thm:rank} and Theorem \ref{thm:rank2} that there exists another local minimum $\hat{W}= \hat{W}=\left\{ \hat{W}_{1},\cdots,\hat{W}_{H}\right\}$, such that $L(\hat{W}) = L(\bar{W})$ and $\rank(\hat{W}_{H}\hat W_{H-1}\cdots\hat{W}_{1})=d_{p}$. Remember that $\hat{R}=\hat{W}_H\hat W_{H-1}\cdots\hat{W}_{1}$. It then follows from Theorem \ref{thm:pertofmatrix} that for any $R$, such that $R$ is a perturbation of $\hat{R}$ and $\rank(R)\le d_p$, we have $R = W_HW_{H-1}\cdots W_1$, where $W_i$ is a perturbation of $\hat W_i$. Therefore, by noticing $\hat{W}$ is a local minimum of \eqref{eq:obj}, we have
$$F(R)= L(W)\geq L(\hat{W}) = F(\hat{R})\ ,$$
which shows that $\hat{R}$ is a local minimum of \eqref{eq:newprob}. \qed

In the proof of Theorem \ref{thm:main2}, we at first show that we just need to consider the case where $X$ is an identity matrix and $Y$ is a diagonal matrix by noticing rotation is invariant under Frobenius norm. Then we show that the local minimum must be a block diagonal and symmetric matrix, and each block term is a projection matrix on the space corresponding to the same eigenvalue of the diagonal matrix $Y$. Finally, we show that those projection matrices must be onto the eigenspace of $Y$ corresponding to the as large as possible eigenvalues, which then shows that the local minimum shares the same function value.

\subsection{Proof of Theorem \ref{thm:main2}}
Let $X=U_1\Sigma_1 V_1^T$ be the SVD decomposition of $X$, where $\Sigma_1$ is a diagonal matrix with full row rank. Then, 
\begin{align*}
F(R) &= \|RU_1 \Sigma_1 V_1^T -Y\|_F^2 
= \|RU_1\Sigma_1 - YV_1\|_F^2 
\\ &= \|(RU_1)(\Sigma_1)_{1:d_1, 1:d_1} - (YV_1)_{1:d_2,1:d_1}\|_F^2 + \mathrm{Const},
\end{align*}
where $\mathrm{Const}$ is a constant in $R$ and $(\cdot)_{t_1:t_2,t_3:t_4}$ is a submatrix of $(\cdot )$, which contains the $t_1$ to $t_2$ row and $t_3$ to $t_4$ column of $(\cdot)$. If $R$ is a local minimum of \eqref{eq:newprob}, then $S=RU_1$ is a local minimum of
\begin{equation}\label{eq:new1prob}
\begin{array}{rl}
\min_{S} & G(S) = \|S\hat{\Sigma}_1 - \hat{Y}\|_F^2\\
s.t. & \rank(S)\le k \ ,
\end{array}
\end{equation}
 where $\hat{\Sigma}_1 := (\Sigma_1)_{1:d_1, 1:d_1}$, $\hat{Y} := (YV_1)_{1:d_2,1:d_1}$ and the difference of objective function values of \eqref{eq:newprob} and \eqref{eq:new1prob} is a constant. Let $\hat{Y} := U_2\Sigma_2 V_2^T$ be the SVD of $\hat{Y}$, then $$G(S)= \|S\hat{\Sigma}_1 - U_2\Sigma_2 V_2^T\|_F^2 = \|U_2^TS\hat{\Sigma}_1V_2 - \hat{\Sigma}_2\|_F^2\ ,$$
and if $S$ is a local minimum of $G(S)$, we have $T:=U_2^TS\hat{\Sigma}_1V_2$ is a local minimum of 
\begin{equation}\label{eq:new2prob}
\begin{array}{rl}
\min_{T} & H(T) = \|T - \Sigma_2\|_F^2\\
s.t. & \rank(T)\le k \ ,
\end{array}
\end{equation}
and the objective function values of \eqref{eq:new1prob} and \eqref{eq:new2prob} are the same at corresponding points. Let $\Sigma_{2}$ have $r$ distinct positive diagonal terms $\lambda_{1}>\cdots>\lambda_{r}\ge0$
with multiplicities $m_{1},\cdots,m_{r}$. Let $T^{*}$ be a local
minimum of \eqref{eq:new2prob}, and 
\[
T^{*}=U^{*}\Sigma^{*}V^{*T}=[U_{S}^{*}U_{N}^{*}]\left[\begin{array}{cc}
\Sigma_{S}^{*} & 0\\
0 & 0
\end{array}\right]\left[\begin{array}{c}
V_{S}^{*T}\\
V_{N}^{*T}
\end{array}\right]
\]
 be the SVD of $T$, where $\Sigma^*_{S}$ are positive singular values.
Let $P_{L}:=U_{S}^{*}\left(U_{S}^{*T}U_{S}^{*}\right)^{-1}U_{S}^{*T}$
and $P_{R}:=V_{S}^{*}(V_{S}^{*T}V_{S}^{*})^{-1}V_{S}^{*T}$ be the
projection matrix to the space spanned by $U_{S}^{*}$ and $V_{S}^{*}$,
respectively. Note that $\left\{ T|P_{L}T=T\right\} \subseteq\left\{ T|rank(T)\le k\right\} $,
thus, $T^{*}$ is also a local minimum of
\begin{align} \label{eq:new3prob}
\min & \|T-\Sigma_{2}\|_{F}^{2}
\\ \nonumber s.t. & P_{L}T=T,
\end{align}
which is a convex problem, and it can be shown by the first order optimality condition that the only local minimum of \eqref{eq:new3prob} is $T^{*}=P_{L}\Sigma_{2}$. Similarly, we have $T^{*}=\Sigma_{2}P_{R}$.
Then, $D:=\Sigma_{2}\Sigma_{2}^{T}$ is a diagonal matrix, with $r$
distinct non-zero diagonal terms $\lambda_{1}^{2}>\cdots>\lambda_{r}^{2}>0$
with multiplicities $m_{1},\cdots,m_{r}$. Therefore,
\begin{align*}
P_{L}DP_{L} &=P_{L}\Sigma_{2}\Sigma_{2}^{T}P_{L}^{T}=T^{*}T^{*T}=\Sigma_{2}P_{R}P_{R}^{T}\Sigma_{2}^{T}
\\ &=\Sigma_{2}P_{R}\Sigma_{2}^{T}=\Sigma_{2}T^{*T}=\Sigma_{2}\Sigma_{2}^{T}P_{L}^{T}=DP_{L}.
\end{align*}
Note that the left hand is a symmetric matrix, thus, $DP_{L}$ is
also a symmetric matrix. Meanwhile, $P_{L}$ is a symmetric matrix,
whereby $P_{L}$ is a $r$-block diagonal matrix with each block corresponding
to the same diagonal terms of $D$. Therefore, $T^{*}=P_{L}\Sigma_{2}$
is also a $r$-block diagonal matrix. 

Let 
\[
T^{*}=\left[\begin{array}{cccc}
T_{1}^{*}\\
 & \ddots\\
 &  & T_{r}^{*}\\
 &  &  & 0
\end{array}\right],
\]
where $T_{i}^{*}$ is a $m_{i}\times m_{i}$ matrix, then $T^{*}T^{*T}=\Sigma_{2}T^{*T}$
implies $T_{i}^{*}T_{i}^{*T}=\lambda_{i}T_{i}^{*T}$. Thus, $T_{i}^{*}$
is a symmetric matrix and $\frac{T_{i}^{*}}{\lambda_{i}}$ is a projection
matrix. Let $rank(T_{i}^{*})=d_{p_{i}}$,
then, $\sum_{i=1}^{r}d_{p_{i}}\le p$ and $tr(T_{i}^{*})=\lambda_{i}d_{p_{i}}$, whereby, 
\begin{align*}
H(T^{*})&=\sum_{i=1}^{r}\|T_{i}^{*}-\lambda_{i}I_{m_{i}}\|_{F}^{2}
\\ &=\sum_{i=1}^{r}tr(T_{i}^{2})-2\lambda_{i}tr(T_{i})+m_{i}\lambda_{i}^{2}
\\ &=\sum_{i=1}^{r}\left(m_{i}-d_{p_{i}}\right)\lambda_{i}^{2}.
\end{align*}
Let $j$ be the largest number that $\sum_{i=1}^{j}m_{i} < d_p$. Then,
it is easy to find that the global minima of \eqref{eq:new3prob} satisfy $d_{p_{i}}=m_{i}$
for $i\le j$, $d_{p_{j+1}}=d_p-\sum_{i=1}^{j}m_{i}$ and $d_{p_i} = 0$ for $i>j+1$ which gives all
of the global minima.

Now, let us show that all local minima must be global minima. As local
minima $T^{*}$ is a block diagonal matrix, thus, we
can assume without loss of generality that both $\Sigma_{2}$ and $T^{*}$ are square matrices, because
the all $0$ rows and columns in $\Sigma_{2}$ and $T$ do not change
anything. Thus, it follows $T_{i}^{*}$ is symmetric that $T^{*}$
is a symmetric matrix. Remember that $\frac{T_{i}^{*}}{\lambda_{i}}$
is a projection matrix, thus the eigenvalues of $T_{i}^{*}$ are either
$0$ or $\lambda_{i}$, whereby 
\[
T^{*}=\sum_{i=1}^{r}\sum_{j=1}^{d_{p_{i}}}\lambda_{i}u_{ij}u_{ij}^{T},
\]
where $u_{ij}$ is the $j$th normalized orthogonal eigen-vector of $T^{*}$
corresponding to eigenvalue $\lambda_{i}$. 

It is easy to see that, at a local minimum, we have $\sum_{i=1}^{r}d_{p_{i}}=d_p$,
otherwise, there is a descent direction by adding a rank 1 matrix
to $T^{*}$ corresponding to one positive eigenvalue. If there exists $i_{1},i_{2}$,
such that $i_{1}<i_{2}$, $d_{p_{i_{1}}}<m_{i_{1}}$, and $d_{p_{i_{2}}}\ge1$,
then, there exists $\bar{u}_{i_{1}}$, such that $\bar{u}_{i_{1}}\perp u_{i_{1}j}$
for $j=1,\cdots d_{p_{i_{1}}}$. Let 
\begin{align*}
T(\theta) := & T^{*}-\lambda_{i_{2}}u_{i_{2}1}u_{i_{2}1}^{T}
+\left(\lambda_{i_{1}}\sin^{2}\theta +\lambda_{i_{2}}\cos^{2}\theta\right) \\
& \left(u_{i_{2}1}\cos\theta + \bar{u}_{i_{1}}\sin\theta\right)(u_{i_{2}1}\cos\theta +\bar{u}_{i_{1}}\sin\theta )^{T}.
\end{align*}

Then, $\rank(T(\theta))=\rank(T^{*})=d_p$, $T(0)=T^{*}$ and $$H(T(\theta))=H(T^{*})+\lambda_{1}^{2}+\lambda_{2}^{2}-\left(\lambda_{1}\sin^{2}\theta+\lambda_{2}\cos^{2}\theta\right)^{2} .$$
It is easy to check that $H(T(\theta))$ is monotonically decreasing with
$\theta$, which gives a descent direction at $T^{*}$, contradicting with
that $T^{*}$ is a local minimum. Therefore, there is no such
$i_{1}$ and $i_{2}$, which shows that $T^{*}$ is a global minimum. \qed



\subsection{Proof of Theorem \ref{thm:main3}}
The statement follows from Theorem \ref{thm:main1} and \ref{thm:main2}.

\section{Conclusion}
We have proven that, even though depth creates a non-convex loss surface, it does not create new bad local minima. Based on this new insight, we have successfully proposed a new simple proof for the fact that all of the local minima of feedforward deep linear neural networks are global minima as a corollary. 

The benefits of this new results are not limited to the simplification of the previous proof. For example, our results apply  to problems beyond square loss. Let us consider the shallow problem (S) $\mini L(R)$  s.t.  $rank(R) \le d_p$, and and the deep parameterization counterpart (D) $\mini L(W_H W_{H-1} \cdots W_1)$.
Our analysis shows that for any function $L$, as long as $L$ satisfies Theorem 3.2, any local minimum of (D) corresponds to a local minimum of (S). This is not limited to when $L$ is least square loss, and this is why we say depth creates no bad local minima.

In addition, our analysis can directly apply to matrix completion unlike previous results.  \citet{ge2016matrix} show that local minima of the symmetric matrix completion problem are global with high probability. This should be able to extend to asymmetric case. Denote
$f(W) := \sum_{i,j \in \Omega} (Y - W_2 W_1)_{i,j}$,
then local minimum of $f(W)$ is global with high probability, where $\Omega$ is the observed entries. Then, our analysis here can directly show that the result can be extended for deep linear parameterization: for
$h(W) := \sum_{i,j \in \Omega} (Y - W_H W_{H-1} \cdots W_1)_{i,j}$,
any local minimum of $h(W)$ is global with high probability.

\section*{Acknowledgements} 
The authors would like to thank Professor Robert M. Freund, Professor Leslie Pack Kaelbling for their generous
support. We also want to thank Cheng Mao for helpful discussions.

\bibliography{LK_DL}
\bibliographystyle{apalike}

\end{document}